\documentclass[12pt]{article}
\usepackage{amsmath}
\usepackage{amssymb}
\usepackage{algorithm}
\usepackage{algorithmic}
\usepackage{tikz}
\usepackage{subcaption} 
\usepackage{comment}

\usepackage{fancyhdr}
\usepackage{graphicx}
\usepackage{color}
\usepackage{comment}

\usepackage[left=2.5cm,right=2.5cm,top=2.5cm,bottom=2.5cm]{geometry}
\begin{document}
\renewcommand{\headrulewidth}{0pt}
\fancyfoot[L]{ JdS2025}
\fancyhead[R]{ }


\begin{center}
{\Large {\sc  Optimal Transport-Based Domain Adaptation for Rotated Linear Regression}}
\bigskip

 \underline{Brian Britos} $^{1}$ \& Mathias Bourel $^{2}$
\bigskip

{\it
$^{1}$ Aix-Marseille Université, France, brian.BRITOS-SIMMARI@etu.univ-amu.fr\\
$^{2}$ Universidad de la República, Uruguay, mbourel@fing.edu.uy}
\end{center}
\bigskip


{\bf R\'esum\'e.} Le transport optimal (OT) s’est révélé efficace pour l’adaptation de domaine (DA) en alignant les distributions entre domaines aux propriétés statistiques différentes. S’appuyant sur l’approche de Courty et al. ~(2016), qui proposaient de projeter les données sources vers le domaine cible afin d’améliorer le transfert de modèles, nous étudions ici un problème supervisé d’adaptation de domaine impliquant des modèles de régression linéaire soumis à des rotations. Ce travail en cours concerne des situations où les domaines source et cible sont liés par une rotation, un cas fréquent dans des contextes tels que l’étalonnage de capteurs ou les changements d’orientation d’images.

Nous montrons que dans $\mathbb{R}^2$, lorsqu’on utilise un coût de type $p$-norme avec $p \geq 2$, le transport optimal permet de retrouver la rotation sous-jacente. À partir de ce résultat théorique, nous proposons un algorithme combinant un regroupement par $K$-moyennes, OT, et une décomposition en valeurs singulières (SVD) pour estimer l’angle de rotation et adapter le modèle de régression. Cette méthode est particulièrement efficace lorsque les données du domaine cible sont limitées, car elle exploite la richesse des données du domaine source pour améliorer la généralisation. Nos contributions offrent ainsi un éclairage à la fois théorique et pratique sur l’utilisation du transport optimal pour l’adaptation de modèles sous des transformations géométriques.

{\bf Mots-cl\'es.} Transport Optimal, Adaptation de Domaine, Régression Linéaire

\medskip

{\bf Abstract.} Optimal Transport (OT) has proven effective for domain adaptation (DA) by aligning distributions across domains with differing statistical properties. Building on the approach of Courty et al. ~ (2016), who mapped source data to the target domain for improved model transfer, we focus on a supervised DA problem involving linear regression models under rotational shifts. This ongoing work considers cases where source and target domains are related by a rotation—common in applications like sensor calibration or image orientation.

We show that in $\mathbb{R}^2$, when using a $p$-norm cost with $p \geq 2$, the optimal transport map recovers the underlying rotation. Based on this, we propose an algorithm that combines $K$-means clustering, OT, and singular value decomposition (SVD) to estimate the rotation angle and adapt the regression model. This method is particularly effective when the target domain is sparsely sampled, leveraging abundant source data for improved generalization. Our contributions offer both theoretical and practical insights into OT-based model adaptation under geometric transformations.


\bigskip\bigskip


\section{Introduction}

Optimal Transport (OT), introduced by Monge (1781) and later formalized by Kantorovich (1942), provides a mathematical framework for efficiently transforming one probability distribution into another. Peyré et al. (2019) offer a detailed overview of its computational aspects, making it a practical tool for various applications. In domain adaptation (DA), OT has proven particularly effective in aligning source and target distributions, especially when labeled data in the target domain are scarce.

In this work, we investigate a supervised DA problem where the source and target domains in $\mathbb{R}^2$ are related by an unknown rotation. We leverage OT to estimate the rotation angle, facilitating the adaptation of a simple linear regression model from the source to the target domain. Our approach combines K-means clustering, OT-based alignment, and a singular value decomposition (SVD) method for robust angle estimation. This framework is particularly useful when target data is limited, allowing the source domain to guide adaptation effectively.

\section{Our method} \label{subsec: R2} 

We begin this section by summarizing two crucial technical tools for our method.

\begin{itemize}
    \item Let $a, b \in \mathbb{R}$ be the coefficients of a line $r$ in the plane $\mathbb{R}^2$. Then, the coefficients of the line obtained by rotating $r$ by an angle $\theta$ around the origin are:
    \begin{equation}
        \tilde{a} = \frac{a \cos\theta + \sin\theta}{\cos\theta - a \sin\theta}, \quad \tilde{b} = b (\cos\theta + a \sin\theta).
    \end{equation}

    \item The algorithm proposed in Arun et al.~(1987) to estimate the rotation angle between two sets in the plane $X^s = \{(x_1^s, y_1^s), \dots, (x_n^s, y_n^s)\}$ and $X^t = \{(x_1^t, y_1^t) , \dots, (x_n^t, y_n^t)\}$.
\end{itemize}

\subsection{Principal theoretical result}
Below we present a novel result concerning a geometric property of optimal transport in the plane. This result will later play a critical role in the design for an algorithm for domain adaptation within the same framework.

\bigskip
\textbf{Proposition:} Let $X^s = \{(x_1^s, y_1^s), \ldots, (x_n^s, y_n^s)\} \subset \mathbb{R}^2$ and $X^t = \{(x_1^t, y_1^t), \ldots, (x_n^t, y_n^t)\} \subset \mathbb{R}^2$ such that $|X^s| = |X^t| = n$, and related by a rotation $R_{\theta}$ of angle $\theta$, that is, each $(x_i^t, y_i^t)$ is of the form $R_{\theta}(x_i^s, y_i^s)$. Let $\mu$ and $\nu$ denote the empirical uniform measures over $X^s$ and $X^t$, respectively, and assume the cost function is the $p$-norm:
\begin{equation*}
    c((x_i^s, y_i^s), (x_j^t, y_j^t)) = \| (x_j^t, y_j^t) - (x_i^s, y_i^s) \|_p, \quad p \geq 2.
\end{equation*}
Then, the optimal transport map $T$ is unique and aligns with the rotation $R_{\theta}$ in the sense that $T((x_i^s, y_i^s)) = R_{\theta}(x_i^s, y_i^s)$ for all $i = 1, \ldots, n$.

\begin{figure}[ht!]
    \centering
    \includegraphics[width=1\linewidth]{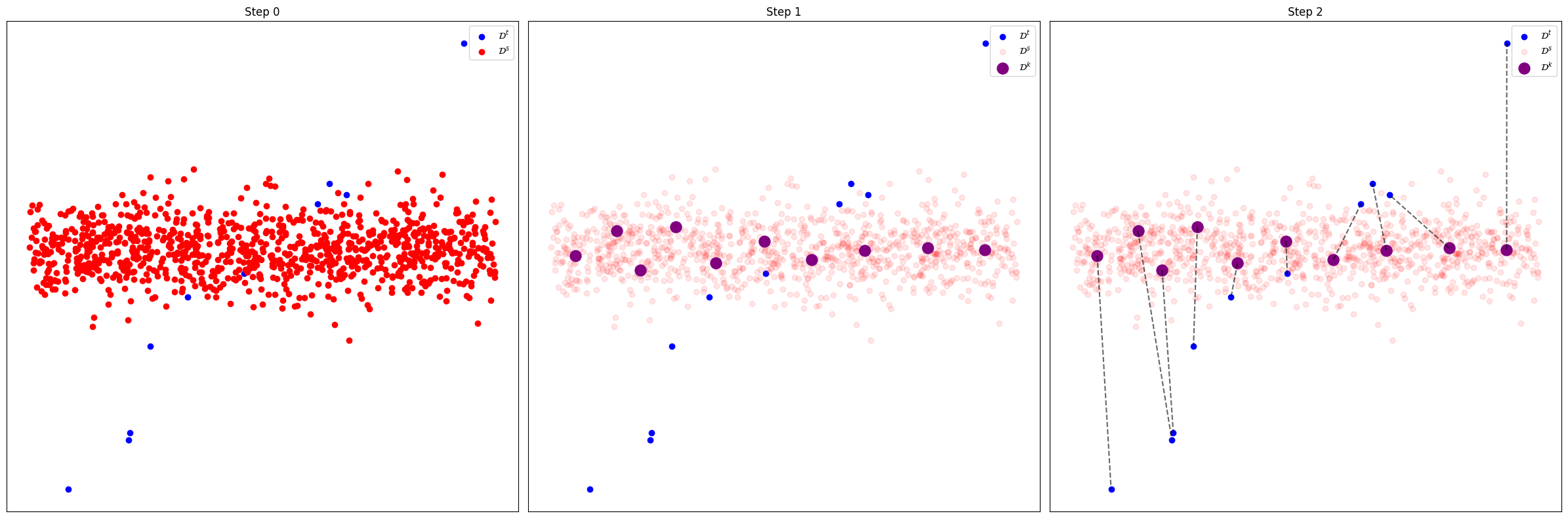}
    \caption{Red points are the source set, blue ones the target set and the purple dots are the $K$-means centroids. Step 0 is the raw data, in Step 1 we apply $K$-means with $K = n_t$ and in Step 3 we find the optimal transport plan.}
    \label{fig:R2}
\end{figure}

\bigskip
\bigskip
Now we shift our attention to the following domain adaptation problem: we have two domains, $\mathcal{D}^s = \{(x_1^s, y_1^s), \ldots, (x_{n_s}^s, y_{n_s}^s)\}$ and $\mathcal{D}^t = \{(x_1^t, y_1^t), \ldots, (x_{n_t}^t, y_{n_t}^t)\}$, in $\mathbb{R}^2$, where $n_s \gg n_t$. We assume that the underlying structure of the two domains are lines related by a unknown rotation $R_{\theta}$, as shown in Step $0$ of Figure \ref{fig:R2}. Here, $\mathcal{D}^s$ serves as a training set and $\mathcal{D}^t$ as a development set, modelling scenarios where training data is abundant but target data is scarce due to rarity, cost, or physical constraints. This setup reflects challenges associated with data drift, common in machine learning (see Sahiner et al.~2023). The goal is to adapt a linear regression model, $y = a_s x + b_s$, trained on $\mathcal{D}^s$, to perform effectively on $\mathcal{D}^t$. The procedure is detailed in the algorithm \ref{alg:rotated_regression}.

Using the above results together with $K-$means we can estimate the angle rotation $\theta$ as follows: first using $K$-means, with $K = n_t$, we construct a set of centroids $\mathcal{D}^k = \{(x_1^k, y_1^k), \ldots, (x_{n_t}^k, y_{n_t}^k)\}$ (shown as purple dots in Step 1 of Figure \ref{fig:R2}). Next, taking uniform measures over these sets and using the $p$-norm as cost function allow us to approximate the assumptions of the previous proposition. Although $\mathcal{D}^t$ and $\mathcal{D}^k$ are not strictly related by a rotation, the optimal transport map $T$ (right of Figure \ref{fig:R2}) approximates the rotation. Finally we apply the proposed algorithm in Arun et al.~(1987) to estimate the rotation angle $\theta$. This procedure is summarized in algorithm \ref{alg:estimacion_del_angulo_SVD}.

\begin{algorithm}[htbp]
\caption{Angle Estimation Using $K$-Means, Optimal Transport, and SVD}
\label{alg:estimacion_del_angulo_SVD}
\begin{algorithmic}
\REQUIRE $\mathcal{D}^s, \mathcal{D}^t$: Source and target domain.
\ENSURE $\hat{\theta}$: Estimated rotation angle.
\STATE \textbf{Step 1:} $\mathcal{D}^k \gets \text{KMeans}(\mathcal{D}_s, \text{n\_clusters} = n_t)$.
\STATE \textbf{Step 2:} $T \gets$ Optimal transport plan obtained with uniform measures $\mu$, $\nu$ and $|| \cdot ||_p$.
\STATE \textbf{Step 3:} Use the optimal transport plan $T$ to sort the sets $\mathcal{D}^t$ and $\mathcal{D}^k$.
\STATE \textbf{Step 4:} $\hat{\theta} \gets$ Estimate rotation angle between $\mathcal{D}^s$ and $\mathcal{D}^k$ using Arun K. S. et al.~(1987).

\RETURN $\hat{\theta}$
\end{algorithmic}
\end{algorithm}

\newpage

\begin{algorithm}[ht!]
\caption{Adapted Linear Regression in $\mathbb{R}^2$}
\label{alg:rotated_regression}
\begin{algorithmic}
\REQUIRE $\mathcal{D}^s, \mathcal{D}^t$: source and target domains; $N$: number of repetitions; $p$: bootstrap proportion.
\ENSURE $a_r, b_r$: adapted regression coefficients.
\STATE Fit linear regression on $\mathcal{D}^s$ to obtain $a_s, b_s$
\FOR{$i = 1$ to $N$}
    \STATE $\mathcal{D}^k \gets$ Sample bootstrap subset $\mathcal{D}^b$ (proportion $p$).
    \STATE $\hat{\theta}$ $\gets$ Apply Algorithm~\ref{alg:estimacion_del_angulo_SVD} to $\mathcal{D}^t$ and $\mathcal{D}^b$.
    \STATE $a_i, b_i \gets$ Apply equation 1 with $a_s, b_s$, and $\hat{\theta}$.
\ENDFOR
\RETURN $a_r = \text{median}(\{a_1, \ldots, a_N \}), b_r = \text{median}(\{b_1, \ldots, b_N \})$
\end{algorithmic}
\end{algorithm}
The motivation for including a bootstrapping step prior to estimating the rotation angle is to ensure the generation of distinct centroids during the application of $K$-means. Additionally, at the end we employ the median instead of the mean, as the procedure has significant outliers.

\section{Simulations}
\label{sec:sim}

In this section we present two simulations to evaluate when the proposed procedure outperforms fitting the regression model solely on the target domain. These tests illustrate its practical advantages and highlight scenarios of significant predictive improvement. We explore two distinct scenarios: in \ref{var_DS}, we vary the source domain size $n_s$ while fixing the angle, noise level, and target domain size $n_t$ whereas in \ref{var_theta_sigma}, we vary the angle and noise level, keeping $n_s$ and $n_t$ fixed. 

\bigskip
The simulations were carried out by drawing points from a straight line and adding Gaussian noise, that means, fixing $a \in \mathbb{R}$ and the noise variance $\sigma > 0$ we generate the data set following $y = ax + \epsilon$ where $\epsilon \sim \mathcal{N}(0, \sigma)$. As we always can apply a rotation to the plane so the source domain is in the $x$-axis, we generate the source domain as $y = \epsilon$ (i.e. $a=0$). On the other hand, fixing an angle $\theta$, we generate the target domain following $y = \arctan(\theta) x + \epsilon$ where the noise variance $\sigma$ is the same for both.


\subsection{Varying the cardinality of $\mathcal{D}^s$} \label{var_DS}
\begin{figure}[h!]
    \centering
    \begin{subfigure}[b]{0.45\textwidth}
        \centering
        \includegraphics[width=\linewidth]{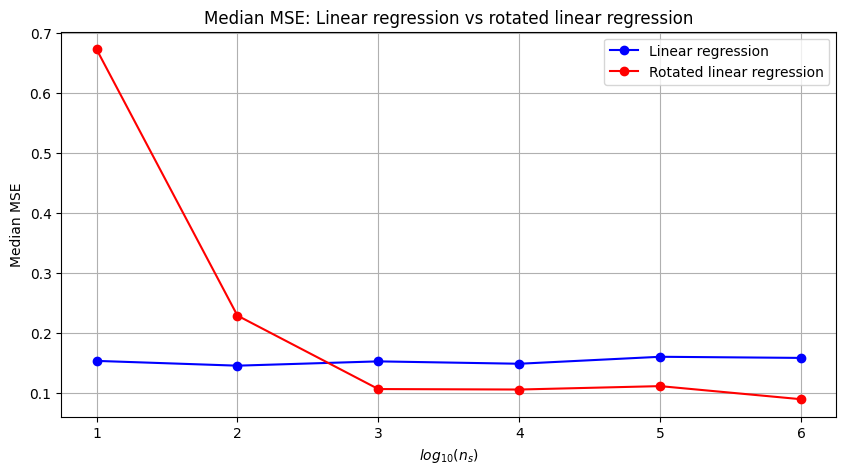}
        \caption{The target domain size $n_t$ is fixed at 10, while the source domain size $n_s$ varies in $\{10, 10^2, 10^3, 10^4, 10^5, 10^6\}$. Blue dots show the median MSE for regression on $\mathcal{D}^t$, and red points show the median MSE using our method.}
        \label{fig:exp_1}
    \end{subfigure}
    \hfill
    \begin{subfigure}[b]{0.45\textwidth}
        \centering
        \includegraphics[width=\linewidth]{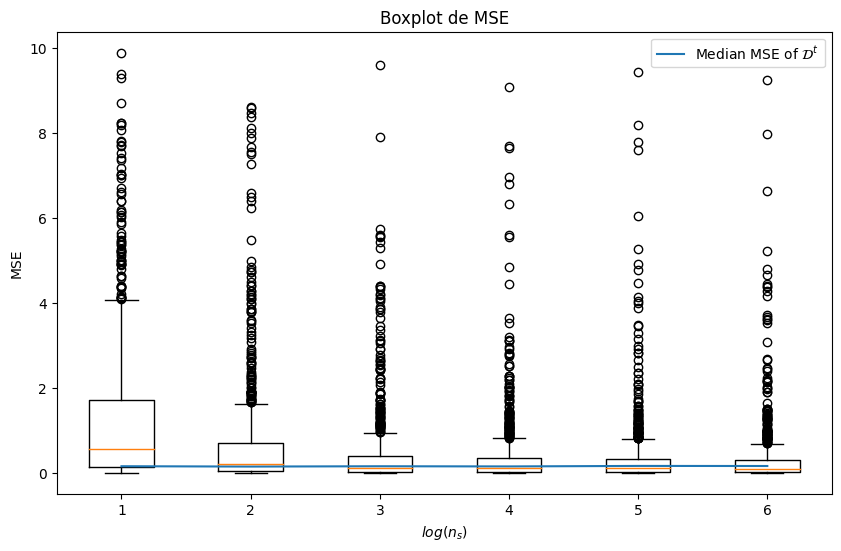}
        \caption{Boxplots of the same experiment as Figure \ref{fig:exp_1}. The blue line represents regression on $\mathcal{D}^t$. Our method shows more outliers but better median performance.}
        \label{fig:exp_1_boplot}
    \end{subfigure}
    \caption{Comparison of the performance of our method against the standard linear regression across different settings of $n_s$.}
    \label{fig:exp_1_combined}
\end{figure}

We simulate data with a fixed rotation angle $\theta = \frac{\pi}{4}$, target domain size $n_t = 10$, noise level $\sigma = 1$ and vary the source domain size $n_s$. For each $n_s$, the procedure is applied $1000$ times, and the median Mean Square Error (MSE) is recorded (Figure \ref{fig:exp_1}). When $n_s$ and $n_t$ are similar (leftmost points), the procedure shows no benefit. However, as the ratio $\frac{n_s}{n_t}$ increases, our method outperforms and exhibits asymptotic behaviour. While the proposed method exhibits a high number of outliers, the median suggests it provides more reliable results in most cases.

\subsection{Varying the angle $\theta$ and noise variance $\sigma$} \label{var_theta_sigma}

\begin{figure}[ht!]
    \centering
    \includegraphics[width=0.6\linewidth]{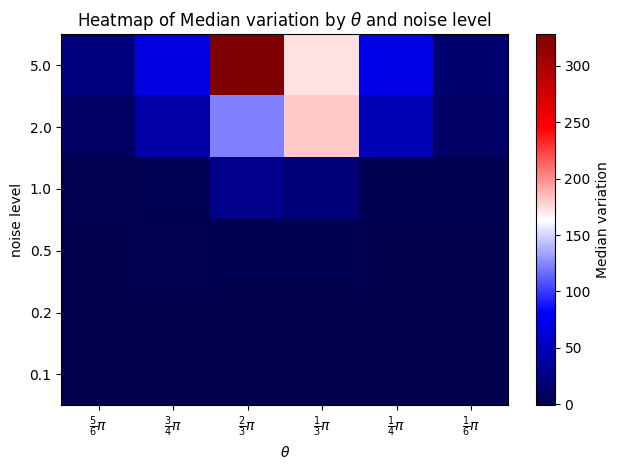}
    \caption{Median variation obtained varying the angle rotation $\theta$ and noise level $\sigma$.}
    \label{fig:exp_2}
\end{figure}

To assess where our method performs best, we fix $n_s = 1000$ and $n_t = 50$, and evaluate variation over the $\theta$–$\sigma$ plane, varying $\theta \in {\frac{5}{6}\pi, \dots, \frac{1}{6}\pi}$ and $\sigma \in {0.1, 0.2, 0.5, 1, 2, 5}$. We defined the variation as $\frac{MSE}{MSE_R}-1$, where $MSE$ is the mean square error of the linear regression obtained using only the target domain $\mathcal{D}^t$ and $MSE_R$ is the mean square error of the regression obtained by our method Figure~\ref{fig:exp_2} shows the median result over 100 runs. Performance improves with higher noise, leveraging the larger source data. Results are also better for $\theta$ near $\frac{\pi}{2}$, likely because linear regression struggles with near-vertical slopes.

\newpage
\section*{Conclusion}

We presents a robust approach to domain adaptation in linear regression using optimal transport, focusing on scenarios where domains are related by rotation. Our results demonstrate that the proposed method achieves better results in $\mathbb{R}^2$, consistently achieving better median performance compared to standard regression methods, even when the target domain size is limited. This highlights the method's potential for practical applications where training data is abundant, but data in the target domain is scarce. Several directions for future improvement and generalization emerge from this study:

\begin{enumerate}
    \item \textbf{Understanding the origin of outliers and improving robustness:}  The proposed method exhibits strong median performance but suffers from significant outliers. A deeper analysis of these outliers will be critical to designing a more robust algorithm that can handle such discrepancies more effectively.
    
    \item \textbf{Generalization to Higher Dimensions:} The extension of the method from \( \mathbb{R}^2 \) to \( \mathbb{R}^n \) using PCA for dimensionality reduction is promising. However, additional research is required to optimize this process.
    
    \item \textbf{Incorporating Translations:} Expanding the framework to incorporate translations will significantly enhance its versatility.
\end{enumerate}

These extensions, combined with the demonstrated effectiveness of our method in \( \mathbb{R}^2 \), provide a foundation for future work aimed at addressing more complex domain adaptation challenges in higher dimensions and diverse scenarios.

\section*{Bibliography}

\noindent Monge G. (1781), Mémoire sur la théorie des déblais et des remblais.

\noindent Kantorovich L. (1942), On the transfer of masses (in russian), \textit{ Dokl. Akad. Nauk SSSR}, , 37, No. 7–8, pp.~227–229.

\noindent Courty N., Flamary R., Tuia D. and Rakotomamonjy A. (2016), Optimal Transport for Domain Adaptation, \textit{IEEE Transactions on pattern analysis and machine intelligence}.

\noindent Arun K. S. , Huang T. S. and Blostein S. D. (1987), Least-squares fitting of two 3-D point sets.

\noindent Peyré G. and Cuturi M. (2019), Computational Optimal Transport, \textit{Foundations and Trends in Machine Learning}.

\end{document}